\documentclass[letterpaper]{article} 
\usepackage[preprint]{aaai2027}  
\usepackage[hyphens]{url}  
\usepackage{graphicx} 
\usepackage{natbib}  
\usepackage{caption} 
\usepackage{booktabs}
\usepackage{amsmath,amssymb}
\title{RoadVGGT: Road-Structure-Aware Feed-Forward Road Surface Reconstruction}
\author{
Han Jiao, Chen Liu\thanks{Project leader.}, Jiakai Sun, Zhanjie Zhang,\\
Mengyuan Yang, Yimeng Li, Mofan Zhou, Kun Zhan, Lei Zhao\corresponding
}
\affiliations{
Zhejiang University\\
Li Auto Inc.\\
csjh@zju.edu.cn, liuchen3@lixiang.com, csjk@zju.edu.cn, cszzj@zju.edu.cn,\\
yangmengyuan@lixiang.com, liyimeng@lixiang.com zhoumofan@lixiang.com, zhankun@lixiang.com,\\
, cszhl@zju.edu.cn
}
\begin{document}
\maketitle
\begin{abstract}
Large-scale road surface reconstruction supports high-definition mapping,
autonomous-driving perception, annotation, and simulation. Existing
road-specialized optimization methods can produce high-quality road
representations, but they typically require per-scene training and
scene-dependent coverage design around the driving trajectory, limiting
scalable reconstruction over newly collected roads. To address these
limitations, we introduce RoadVGGT, a
road-structure-aware feed-forward framework that reconstructs compact
Gaussian road surfaces without test-time per-scene optimization. RoadVGGT
uses a geometric foundation model to exploit multi-view images together with
provided pose and depth observations, and predicts dense pixel-aligned
Gaussian attributes through a learned Gaussian head. To make these dense
predictions usable for large road surfaces, we align them into a consistent
metric world coordinate system and fuse redundant Gaussians on the
road-aligned XY plane through confidence-weighted grid fusion. Category-aware grouping and
road--sidewalk junction protection further control fusion around vulnerable
road structures. The resulting representation supports RGB and semantic
bird's-eye-view maps, elevation estimation, and novel view synthesis.
RoadVGGT eliminates the need for per-scene optimization in prior methods,
reconstructs complete road surfaces with a compact Gaussian representation,
and improves image quality, semantic mapping, and elevation accuracy. Extensive
experiments demonstrate the potential of geometric foundation models for
scalable feed-forward road surface reconstruction.
\end{abstract}
\section{Introduction}
Road surface reconstruction aims to recover the appearance, semantics, and
geometry of roads over extended trajectories. Such reconstructions can
support high-definition mapping, road-marking perception, automatic
annotation, simulation, and bird's-eye-view (BEV) reasoning. Unlike generic
novel view synthesis, road reconstruction must represent large smooth
regions while retaining thin markings, subtle elevation changes, and
boundaries between nearby surfaces. A useful road representation should also
support several downstream products, including RGB and semantic BEV maps,
elevation maps, and novel view synthesis.

Recent neural radiance field and Gaussian Splatting methods
\citep{snerf,drivinggaussian,autosplat,s3gaussian,Zhou_2024_CVPR} have substantially
improved outdoor and driving-scene reconstruction. In particular, RoGS
\citep{rogs}
demonstrates that flattened Gaussian surfels arranged over a road plane form
an effective representation for road appearance, semantics, and elevation.
However, these optimization-based road-scene reconstruction methods optimize a
new representation for each target scene. Their reconstruction quality comes with iterative per-scene
training, and road-specific pipelines often require scene-dependent coverage
or hyperparameter choices around the driving trajectory. This limits rapid
deployment and map updates over continuously collected or previously unseen
roads.
Geometric foundation models create an opportunity to replace scene-specific
optimization with direct inference. DUSt3R \citep{dust3r} reformulates
reconstruction as pointmap regression; CUT3R \citep{cut3r} extends learned
geometry to continuous observations; and VGGT \citep{vggt} directly predicts
cameras, depth, pointmaps, and tracks from many views. OmniVGGT
\citep{omnivggt} further supports available camera and depth observations. This
capability is particularly suitable for road-mapping platforms because pose
and depth measurements are often available together with multi-view images in
driving scenes.
Nevertheless, geometric prediction alone does not provide a compact,
renderable road surface representation. Point-based geometry lacks the
appearance and rasterization properties needed for high-quality road map
rendering, whereas Gaussian primitives can attach color, opacity, semantic,
and geometric attributes to the predicted road structure. Therefore, a natural
solution is to attach a Gaussian prediction head to the feed-forward
geometric backbone.

However, directly applying this feed-forward, pixel-aligned Gaussian
prediction head at every input pixel
produces one Gaussian per pixel.
For long multi-camera road sequences, the resulting representation scales
with both image resolution and view count. This creates extensive redundancy
over locally smooth road surfaces and increases storage and
rendering cost. Generic feed-forward Gaussian methods
\citep{pixelsplat,mvsplat,noposplat,smart2024splatt3r} commonly rely on dense
pixel-aligned predictions.
AnySplat \citep{anysplat} addresses the resulting redundancy through
voxel-guided 3D fusion. This generic volumetric strategy is designed for
general novel view synthesis and does not exploit the fact that road surfaces
are locally close to a ground-aligned manifold. It also lacks a metric
road-plane grid whose cell size has the same physical meaning across
trajectories of different lengths. Recent token-based approaches reduce
redundancy before Gaussian decoding. TokenSplat \citep{tokensplat} fuses
spatially corresponding image tokens, whereas C3G \citep{c3g} and GlobalSplat
\citep{globalsplat} decode a fixed Gaussian budget from global learnable
tokens. Such fixed-budget designs are poorly matched to road surfaces whose
spatial extent varies with trajectory length.

Road structure also complicates aggressive fusion. Uniform fusion can blur
thin markings or mix observations from nearby but distinct surfaces.
Road and sidewalk Gaussians, for example, may occupy the same ground-plane
cell while belonging to adjacent surfaces with different semantics,
appearances, or local elevations. Compression must therefore distinguish
smooth regions from structures that should retain finer support.

To address these issues, we introduce RoadVGGT, a road-structure-aware feed-forward framework for
large-scale road surface reconstruction. RoadVGGT uses OmniVGGT as its
geometric backbone and predicts dense Gaussian attributes from multi-view
features. It then converts the dense predictions into a compact road
representation through confidence-weighted XY grid fusion. Category-aware
grouping preserves selected vulnerable structures, while road--sidewalk
junction protection prevents destructive cross-surface fusion. The
resulting Gaussian representation supports multiple downstream tasks without
test-time scene-specific optimization.
\begin{figure*}[t]
\centering
\includegraphics[width=\textwidth]{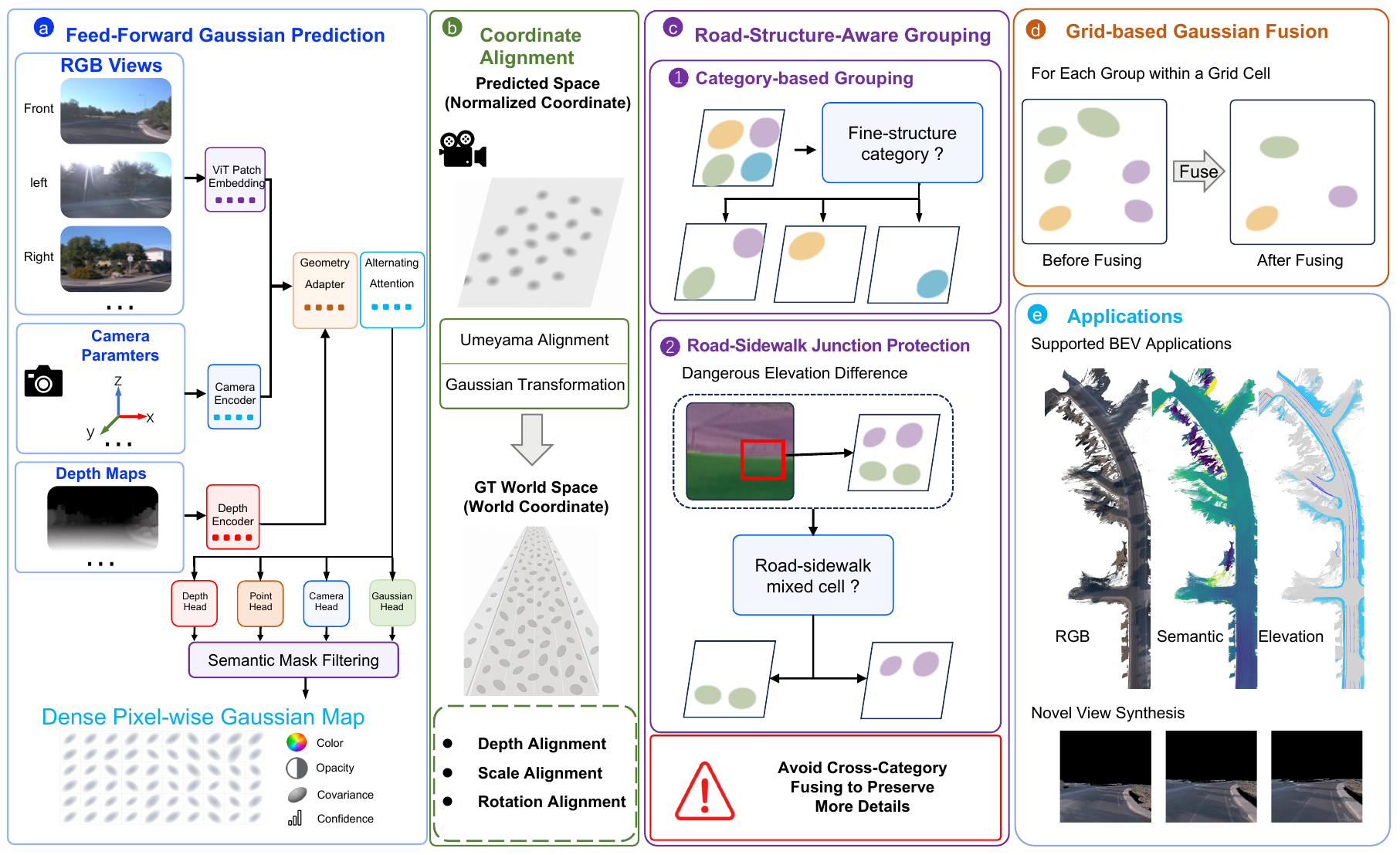}
\vspace{-5mm}
\caption{
Overview of RoadVGGT. Given multi-view road images with provided camera poses,
depth observations, and semantic segmentation maps, RoadVGGT predicts dense Gaussian attributes, aligns
them to a metric road-plane coordinate system, and applies road-structure-aware
grid fusion to produce a compact Gaussian road representation. The resulting Gaussian representation supports BEV RGB, semantic, and elevation outputs, as well as camera-view
rendering and novel view synthesis.
}
\label{fig:overview}
\vspace{-4mm}
\end{figure*}

Our contributions are as follows:
\begin{itemize}
    \item We present RoadVGGT, a road-structure-aware feed-forward
    framework that reconstructs compact Gaussian road surfaces without
    test-time per-scene optimization and scene-specific hyperparameter
    tuning.
    \item We introduce road-plane grid fusion to fuse dense pixel-wise
    Gaussian predictions, reducing storage and rendering cost while
    preserving reconstruction quality.
    \item We introduce category-aware grouping and road--sidewalk junction
    protection to preserve vulnerable road structures under Gaussian
    fusion.
\end{itemize}
\section{Related Work}
\subsection{Road and Street-Scene Reconstruction}
Neural Radiance Fields (NeRF) \citep{nerf} represent a scene as a continuous
radiance field and synthesize novel views through volume rendering. 3D
Gaussian Splatting (3DGS) \citep{3dgs} instead uses explicit anisotropic
Gaussians to represent a scene. Both representations have been extended to
driving scenes. S-NeRF \citep{snerf} adapts NeRF to large-scale street views with
limited overlap between surround-view cameras. DrivingGaussian
\citep{drivinggaussian} decomposes a dynamic scene into static and
object-level Gaussian components; S3Gaussian \citep{s3gaussian} learns
dynamic street Gaussians with self-supervision; HUGS \citep{Zhou_2024_CVPR} jointly
optimizes geometry, appearance, semantics, and motion; and AutoSplat
\citep{autosplat} introduces road and sky constraints. Most of these methods
reconstruct complete driving environments for rendering or simulation and
generally optimize a separate representation for each scene.
Road surface reconstruction instead targets a long, predominantly
surface-like region while retaining map-relevant appearance, semantics, and
elevation. RoMe \citep{rome} represents large road surfaces with an explicit
mesh and learned elevation, while EMIE-MAP \citep{emie-map} combines an
explicit mesh with implicit color encoding to handle appearance differences
among surround-view cameras. RoGS \citep{rogs} replaces the mesh with
meshgrid-distributed 2D Gaussian surfels for road rendering and mapping. Despite their
different representations, these road-specific methods rely on iterative
per-scene optimization. RoadVGGT instead directly infers a compact Gaussian
road surface without optimizing a new representation for every scene.
\subsection{Feed-Forward Geometric Foundation Models}
Learned geometric models have progressively reduced the need for classical
scene-specific optimization. DUSt3R \citep{dust3r} predicts pointmaps from
unconstrained image pairs. CUT3R \citep{cut3r} maintains a persistent state
for continuous 3D reconstruction. VGGT \citep{vggt} scales direct geometric
prediction to many views and jointly estimates cameras, depth, pointmaps, and
tracks. Depth Anything 3 \citep{depthanything3} recovers spatially consistent
geometry from an arbitrary number of views through unified depth--ray
prediction, with or without known camera poses. OmniVGGT \citep{omnivggt}
generalizes multi-view geometric prediction to arbitrary available geometric
observations, including camera poses and depth.
RoadVGGT leverages the strong geometric reasoning capability of such models
to extract multi-view features and geometric predictions from images
together with provided camera poses and depth observations, and further
converts them into a compact road-surface representation through a
dedicated Gaussian prediction head and road-structure-aware fusion.
\subsection{Feed-Forward Gaussian Reconstruction}
Feed-forward Gaussian methods directly predict scene representations from
input images. pixelSplat \citep{pixelsplat} and MVSplat \citep{mvsplat} infer
pixel-aligned Gaussians from sparse posed views; NoPoSplat
\citep{noposplat} extends this setting to unposed images; and Splatt3R
\citep{smart2024splatt3r} predicts Gaussian attributes on foundation-model pointmaps.
However, dense pixel-aligned prediction produces substantial cross-view
redundancy. Token-based methods instead reduce redundancy before Gaussian
decoding. TokenSplat \citep{tokensplat} aligns and fuses spatially
corresponding image tokens, with the output determined by the fused tokens.
By contrast, C3G \citep{c3g} and GlobalSplat \citep{globalsplat} decode a
fixed Gaussian budget from global learnable tokens. This fixed capacity does
not adapt naturally to road segments with widely varying physical extent. AnySplat
\citep{anysplat} follows a spatial fusion strategy instead, consolidating
decoded Gaussians in a 3D voxel grid. Because it is designed for generic
novel view synthesis, this volumetric partition does not exploit
ground-aligned road structure or preserve road details during fusion.
Feed-forward Gaussian reconstruction has also been extended to driving-scene
reconstruction.
ADGaussian \citep{adgaussian} combines image and sparse-depth features for
generalizable street view synthesis; GGS \citep{ggs} focuses on
large-viewpoint lane switching; and ReconDrive \citep{recondrive} and DGGT
\citep{dggt} both adopt 4D Gaussians to model dynamic driving scenes.
RoadVGGT addresses a different
target: an explicit static road-surface map that supports appearance,
semantics, and elevation over long trajectories. It fuses dense predictions
on a metric road-plane grid, allocating Gaussian capacity according to
physical road coverage so that road segments of different lengths can all be
modeled at fine spatial granularity. In addition, road-structure-aware grouping further
preserves road-surface details during fusion.
\section{Method}
\subsection{Overview and Problem Formulation}
Given \(V\) multi-view road observations, we denote the inputs by
\[
\mathcal{X}=\{I_v,K_v,T_v,D_v,S_v\}_{v=1}^{V},
\]
where \(I_v\), \(K_v\), \(T_v\), \(D_v\), and \(S_v\) are the RGB image,
camera intrinsics, camera pose, depth observation, and
pixel-level semantic segmentation map of view \(v\), respectively. The
segmentation maps are obtained from a fine-tuned Mask2Former \citep{mask2former}
model and are
system inputs rather than outputs of the Gaussian head.
Our goal is to reconstruct a compact road Gaussian representation
\[
\mathcal{G}=\{g_i=(\boldsymbol{\mu}_i,\mathbf{c}_i,\alpha_i,
\mathbf{s}_i,\mathbf{r}_i)\}_{i=1}^{N},
\]
where \(\boldsymbol{\mu}_i\), \(\mathbf{c}_i\), \(\alpha_i\),
\(\mathbf{s}_i\), and \(\mathbf{r}_i\) denote position, appearance, opacity,
scale, and rotation. Each Gaussian is associated with an external discrete
semantic label \(\ell_i\) inherited from its source pixel. The label is
auxiliary metadata and is not predicted as a Gaussian attribute.
RoadVGGT reconstructs \(\mathcal{G}\) without test-time scene-specific
optimization. The resulting representation supports BEV generation and
multiple downstream tasks.
Figure~\ref{fig:overview} illustrates the RoadVGGT pipeline.
\subsection{Feed-Forward Gaussian Prediction}
Let \(\Phi\) denote the geometric foundation backbone. We instantiate it with
OmniVGGT because it exploits the provided camera poses and depth observations
together with multi-view images. Given \(\mathcal{X}\), the backbone produces
dense depth estimates, camera predictions, and multi-scale features:
\[
\{\hat D_v,\hat T_v,\mathbf{F}_v\}_{v=1}^{V}
=\Phi(\mathcal{X}).
\]
Here, \(\hat D_v\), \(\hat T_v\), and \(\mathbf{F}_v\) denote the predicted
depth, camera pose, and multi-scale feature tokens of view \(v\), respectively.
On top of this backbone, RoadVGGT introduces a DPT-style Gaussian head
\(\Psi_{\mathrm{G}}\) for dense road appearance modeling. The head decodes the
multi-scale feature tokens and predicts dense Gaussian attributes and an
additional fusion confidence for each input pixel:
\[
(\hat g_{v,p},\rho_{v,p})=\Psi_{\mathrm{G}}(\mathbf{F}_v)_p.
\]
Here, \(p\) indexes a pixel in view \(v\), \(\hat g_{v,p}\) contains
appearance, opacity, scale, and rotation, and \(\rho_{v,p}\) estimates the
reliability of the prediction during fusion. The Gaussian center is not
decoded by the head; instead, it is recovered by back-projecting the
predicted depth of the corresponding pixel. We use
flattened 2D Gaussians by fixing the \(z\)-axis scale component to zero,
matching the locally surface-like structure of roads.
To place all predicted Gaussians in a shared metric world coordinate system,
we align the predicted camera trajectory with the input trajectory using a
least-squares similarity transform. Given predicted and input camera centers
\(\mathbf{o}^{pred}_v\) and \(\mathbf{o}^{in}_v\), we estimate a scale
\(s\), a rotation matrix \(R\), and a translation \(\mathbf{t}\):
\[
(\hat s,\hat R,\hat{\mathbf{t}})
=\arg\min_{s,R,\mathbf{t}}\sum_v
\left\|\mathbf{o}^{in}_v-
\left(sR\mathbf{o}^{pred}_v+\mathbf{t}\right)\right\|_2^2.
\]
The resulting alignment is then applied to Gaussian geometry in three parts.
First, predicted depth is scaled by \(\hat s\) and unprojected with the input
camera intrinsics and camera-to-world poses to obtain Gaussian centers in a
shared metric world coordinate system. Second, Gaussian scales are multiplied
by \(\hat s\). Third, Gaussian rotations are transformed by \(\hat R\). This
aligned coordinate system provides two properties for subsequent grid fusion:
the horizontal \(XY\) plane serves as a natural road-plane reference, while
the metric scale makes grid cells physically meaningful and comparable across
both short clips and long road segments.
\begin{figure*}[!t]
\centering
\includegraphics[width=\textwidth]{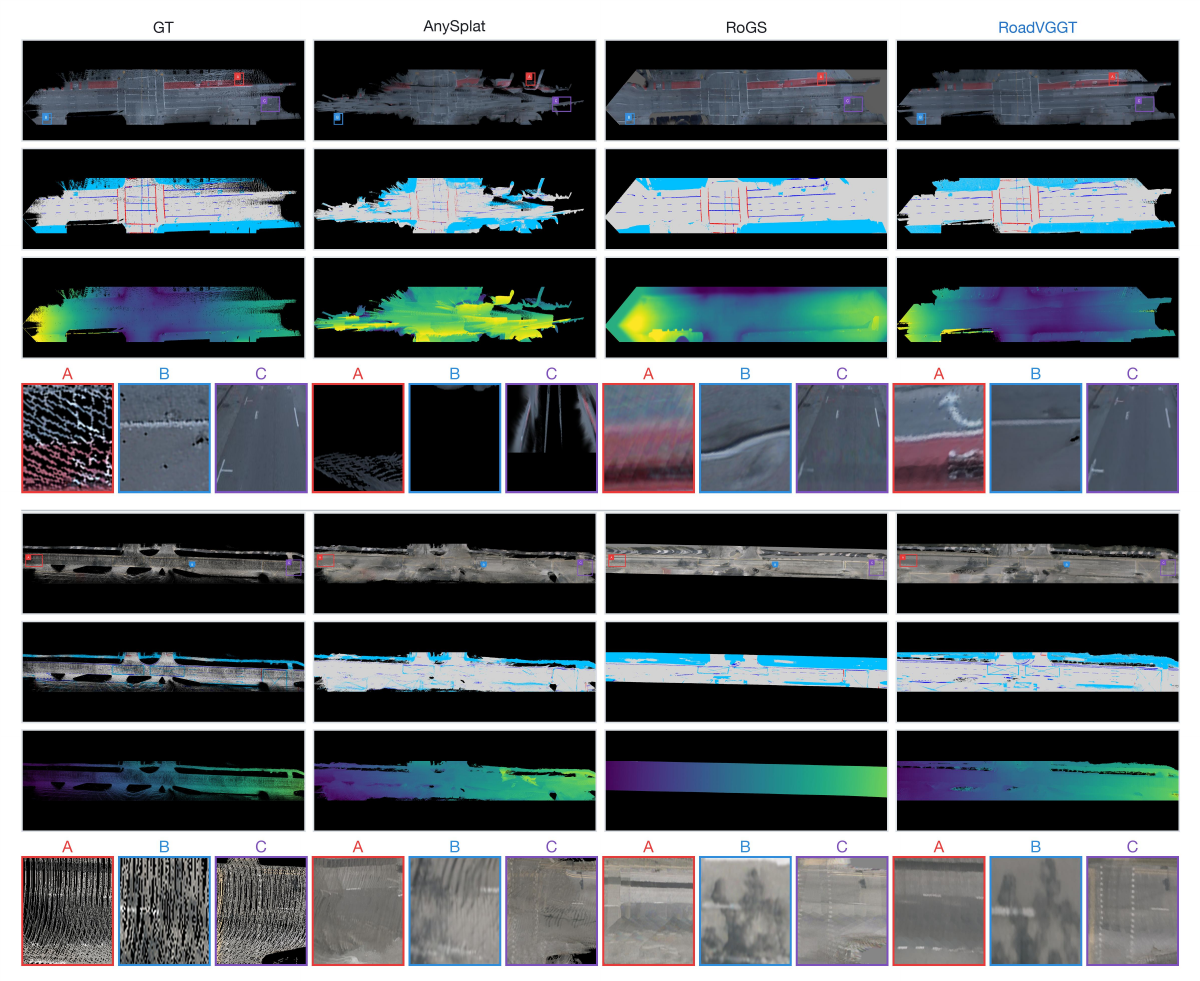}
\vspace{-6mm}
\caption{Qualitative comparison of road-surface reconstruction on Waymo and
zero-shot nuScenes. Columns show ground truth, AnySplat, RoGS, and RoadVGGT.
For each example, the first three rows present RGB BEV, semantic BEV, and elevation.
In the Waymo example, A and B are enlarged BEV regions and C is a camera-view
comparison; in the nuScenes example, A--C are enlarged BEV regions.
\textbf{Note}: within each method column, the enlarged views follow A, B, and C
from left to right.}
\label{fig:qualitative}
\vspace{-4mm}
\end{figure*}

\begin{table*}[t]
\centering
\small
\begin{tabular*}{\textwidth}{@{\extracolsep{\fill}}lccccccccc}
\toprule
Method & Test Opt. & PSNR$\uparrow$ & SSIM$\uparrow$ & LPIPS$\downarrow$ &
mIoU$\uparrow$ & Z-RMSE$\downarrow$ & Storage (MB)$\downarrow$ & Time (s)$\downarrow$ & Render (s)$\downarrow$ \\
\midrule
\multicolumn{10}{l}{\textit{(a) Waymo}} \\
AnySplat & No & 24.09 & 0.8307 & 0.1829 & 0.2768 & 4.8173 & 712.47 & \textbf{260.32} & 3.13 \\ 
RoGS & Yes & 24.81 & 0.8301 & 0.1943 & 0.4361 & 0.2635 & 178.67 & 362.94 & 1.05 \\ 
RoadVGGT & No & \textbf{25.71} & \textbf{0.8480} & \textbf{0.1769} &
\textbf{0.4692} & \textbf{0.2312} & \textbf{112.81} & 263.04 & \textbf{0.84} \\ 
\midrule
\multicolumn{10}{l}{\textit{(b) nuScenes}} \\
AnySplat & No & 21.17 & 0.8935 & 0.2935 & 0.2859 & 0.5595 &
880.45 & 205.92 & 3.65 \\ 
RoGS & Yes & 22.55 & 0.9070 & 0.2516 & 0.4199 & 0.2404 &
85.23 & \textbf{197.80} & 0.67 \\ 
RoadVGGT & No & \textbf{23.80} & \textbf{0.9205} & \textbf{0.2422} &
\textbf{0.4722} & \textbf{0.2241} & \textbf{78.10} & 229.64 & \textbf{0.63} \\ 
\bottomrule
\end{tabular*}
\vspace{-2mm}
\caption{Average reconstruction quality, representation size, runtime, and rendering time
on Waymo and nuScenes. RoadVGGT is trained only on
Waymo and evaluated zero-shot on nuScenes.}
\label{tab:quantitative-comparison}
\vspace{-4mm}
\end{table*}

\subsection{Grid-Based Gaussian Fusion}
Directly retaining all predictions yields approximately \(VHW\) Gaussians,
where \(H\) and \(W\) are the input image height and width.
Before fusion, semantic labels are used to retain road-surface-related
Gaussian candidates and discard background ones. We then exploit the
local smoothness of roads by grouping the retained Gaussians on an XY-aligned
grid. We use a 2D road-plane grid rather than a generic 3D voxel grid because
the road-surface Gaussians primarily vary over the horizontal plane and occupy
a relatively limited range in height. Within this 2D grouping, the larger
number of accurately predicted near-plane Gaussians can also suppress a small
number of off-plane outliers caused by geometry prediction inaccuracies during
fusion.\linebreak Let \(\boldsymbol{\mu}_i=(x_i,y_i,z_i)\) denote the metric center of
Gaussian \(i\), and let \(x_{\min}\) and \(y_{\min}\) denote the minimum XY
coordinates over the retained Gaussians. For grid resolution \(r\), Gaussian
\(i\) is assigned to the
XY-grid index
\[
u_i=\left\lfloor\frac{x_i-x_{\min}}{r}\right\rfloor,\qquad
v_i=\left\lfloor\frac{y_i-y_{\min}}{r}\right\rfloor.
\]
Let \(k\) index an occupied XY cell, and let \(\mathcal{C}_k\) denote the
Gaussians assigned to it. We compute confidence-weighted fusion within each
occupied XY cell:
\[
w_i=\frac{\exp(\rho_i)}
{\sum_{j\in\mathcal{C}_k}\exp(\rho_j)},\quad
\bar a_k=\sum_{i\in\mathcal{C}_k}w_i a_i,\quad i\in\mathcal{C}_k.
\]
Here, \(\rho_i\) is the confidence associated with Gaussian \(i\), and
\(\bar a_k\) denotes the fused value of a non-semantic Gaussian attribute
\(a_i\) in cell \(k\).
This produces one fused Gaussian \(\bar g_k\) per occupied cell.
External semantic labels are discrete and are propagated by majority voting
over semantic classes \(m\), where \(\mathbf{1}[\cdot]\) is the indicator
function:
\[
\bar\ell_k=\arg\max_m\sum_{i\in\mathcal{C}_k}
\mathbf{1}[\ell_i=m].
\]
This operation converts redundant pixel-wise predictions into a compact
cell-level road representation.
\subsection{Road-Structure-Aware Grouping}
Uniform fusion works well on smooth regions but can blur thin markings and
road-structure boundaries. This issue mainly appears in two cases. First, fine
semantic structures such as markings or crosswalks may share an XY cell with
ordinary road-surface Gaussians and become absorbed by the dominant surface.
Second, road and sidewalk Gaussians may co-occur near a boundary and should
not be averaged into a single surface. RoadVGGT addresses these cases by
forming structure-compatible groups before fusion. In other words, Gaussians are fused only
within the same group, preserving compactness while avoiding destructive
cross-structure averaging.
\paragraph{Category-aware grouping.}
To handle the first case, let \(\mathcal{L}_{fine}\) denote the fine-structure
categories and let \(h_i^{cat}\) be the category-group identity of Gaussian
\(i\). Gaussians whose labels are in \(\mathcal{L}_{fine}\) keep their
semantic label as the group identity, while all remaining Gaussians are
assigned to a shared base group for this partition:
\[
h_i^{cat}=
\begin{cases}
\ell_i, & \ell_i\in\mathcal{L}_{fine},\\
\mathrm{base}, & \text{otherwise}.
\end{cases}
\]
Within each XY cell, this prevents fine structures from being fused into the
ordinary surface group, while still allowing Gaussians from the same fine
category to be fused together.
\paragraph{Road--sidewalk junction protection.}
To handle the second case, we identify cells where road and sidewalk
Gaussians co-occur:
\[
J_k=\mathbf{1}\left[
\exists i,j\in\mathcal{C}_k:
\ell_i=\mathrm{road}\land\ell_j=\mathrm{sidewalk}\right].
\]
In these mixed cells, road and sidewalk observations are assigned to separate
junction groups, while all remaining Gaussians stay in the junction base
group, where \(k(i)\) denotes the XY cell containing Gaussian \(i\):
\[
h_i^{junc}=
\begin{cases}
\ell_i, & J_{k(i)}=1\ \text{and}\ \ell_i\in\{\mathrm{road},\mathrm{sidewalk}\},\\
\mathrm{base}, & \text{otherwise}.
\end{cases}
\]
The final group identity is the ordered pair of category and junction labels,
\[
h_i=(h_i^{cat},h_i^{junc}).
\]
Thus, confidence-weighted fusion is applied independently within each
non-empty subgroup, so two Gaussians are fused together only when both group
identities match:
\[
\mathcal{C}_{k,h}=\{i\in\mathcal{C}_k\mid h_i=h\},
\]
and the outputs of all subgroups are concatenated to form the final Gaussian
road representation.
\subsection{Training and Inference}
We freeze the camera, depth, and point prediction components and optimize the
Gaussian head through differentiable rendering. The training objective is
defined on rendered camera-view images and combines an L1 reconstruction loss
with a perceptual LPIPS loss, where \(\lambda\) denotes the LPIPS weight:
\[
\mathcal{L}=\mathcal{L}_{L1}+\lambda\mathcal{L}_{LPIPS}.
\]
Here, \(\mathcal{L}_{L1}\) and \(\mathcal{L}_{LPIPS}\) are computed between
rendered and ground-truth camera-view RGB images.
Grid fusion is applied before rendering during training, so the Gaussian head
learns attributes that remain effective after the same fusion operation used at
inference time.
At inference time, a long road sequence is divided into non-overlapping chunks
of consecutive timestamps, each containing synchronized images from all
cameras. Each chunk is processed independently to predict Gaussians, which are
transformed into the shared metric world coordinate system. The Gaussians from
all chunks are then combined, followed by a single global road-structure-aware
grid fusion.
\section{Experiments}
\subsection{Experimental Setup}
\paragraph{Datasets and protocol.}
We conduct experiments primarily on the Waymo Open Dataset \citep{waymo}
perception benchmark, which contains real-world driving logs
with synchronized multi-camera images, LiDAR measurements, and camera
calibration. Each Waymo segment provides images from five cameras. RoadVGGT
is trained on the Waymo training split and evaluated on the
official Waymo test split. To test cross-dataset
generalization, we also evaluate on nuScenes \citep{nuscenes}, a large-scale
driving dataset with 1000 20-second scenes captured by a
six-camera surround-view system and LiDAR; no nuScenes scene is used for
training. For both datasets, we obtain pixel-level semantic segmentation maps
using a fine-tuned Mask2Former \citep{mask2former} model. These semantic maps
are used for road-surface filtering and structure-aware grouping. We report
PSNR, SSIM, and LPIPS for rendered image quality, mIoU for BEV semantics, and
the root mean squared error between predicted and ground-truth road elevations
(Z-RMSE) after median-based Z-offset alignment. Because different methods may
reconstruct different spatial extents, we compute all quality metrics only
over their common valid region, defined by the intersection of the
reconstruction masks of the compared methods.
\paragraph{Implementation details.}
We initialize RoadVGGT from an OmniVGGT checkpoint, freeze the geometric
backbone, and optimize only the newly introduced Gaussian head on the Waymo
training split. Training uses 20k sampled instances per epoch, with 16 input
images per instance. We use AdamW with a learning rate of
\(4\times10^{-5}\), weight decay \(0.01\), 100 warmup steps, and a cosine
learning-rate schedule. The LPIPS term is linearly warmed up during the first
100 training steps and then kept at a weight of \(0.05\). We train for 8
epochs on 8 NVIDIA A100 GPUs. The default road-plane grid resolution is
\(r=0.05\) m. During inference, chunks contain 6 consecutive Waymo timestamps
or 5 consecutive nuScenes timestamps. Additional implementation details, per-scene results, and extended evaluations are provided in the supplementary material.
\paragraph{Baseline.}
We compare with AnySplat \citep{anysplat}, a feed-forward Gaussian
reconstruction baseline that relies on generic voxel-grid fusion, and RoGS \citep{rogs}, a representative
optimization-based baseline. RoadVGGT and AnySplat perform direct inference
for each scene, whereas RoGS trains a scene-specific representation. 
\subsection{Quantitative Comparison}
\begin{table*}[t]
\centering
\small
\begin{tabular*}{\textwidth}{@{\extracolsep{\fill}}lccccccc}
\toprule
Variant & PSNR$\uparrow$ & SSIM$\uparrow$ & LPIPS$\downarrow$ &
mIoU$\uparrow$ & Z-RMSE$\downarrow$ & Storage (MB)$\downarrow$ & Time (s)$\downarrow$ \\
\midrule
\multicolumn{8}{l}{\textit{(a) Dense Prediction and Grid Resolution}} \\
Dense & 24.22 & 0.7858 & 0.1977 & 0.4176 & 0.4808 & 1864.67 & 273.64 \\ 
0.02 m & 25.22 & 0.8287 & \textbf{0.1746} & \textbf{0.4850} & 0.4297 & 494.09 & 272.70 \\ 
0.05 m & \textbf{25.71} & \textbf{0.8480} & 0.1769 & 0.4692 & 0.2312 & 112.81 & 263.04 \\ 
0.10 m & 25.46 & 0.8455 & 0.1803 & 0.2501 & \textbf{0.2040} & 33.07 & 253.02 \\ 
0.20 m & 23.40 & 0.7649 & 0.2163 & 0.1812 & 0.2732 & 8.91 & 251.88 \\ 
\midrule
\multicolumn{8}{l}{\textit{(b) Gaussian Parameterization}} \\
3D Gaussian & 24.47 & 0.8086 & 0.1789 & 0.3773 & 0.3112 & 112.81 & 263.88 \\ 
2D Gaussian & \textbf{25.71} & \textbf{0.8480} & \textbf{0.1769} &
\textbf{0.4692} & \textbf{0.2312} & 112.81 & 263.04 \\ 
\midrule
\multicolumn{8}{l}{\textit{(c) Road-Structure-Aware Grouping}} \\
Full Model & \textbf{25.71} & \textbf{0.8480} & \textbf{0.1769} & \textbf{0.4692} &
\textbf{0.2312} & 112.81 & 263.04 \\ 
w/o Junc. & 25.69 & 0.8464 & 0.1821 & 0.4658 &
0.2638 & 102.99 & 258.58 \\ 
w/o Class+Junc. & 25.51 & 0.8385 & 0.1910 & 0.4499 &
0.2606 & 96.08 & 245.59 \\ 
\bottomrule
\end{tabular*}
\vspace{-2mm}
\caption{Ablation studies on Waymo. The three groups evaluate grid resolution,
Gaussian parameterization, and road-structure-aware grouping, respectively,
together with their runtime.}
\label{tab:ablation-summary}
\vspace{-2mm}
\end{table*}

\begin{figure*}[t]
\centering
{\footnotesize
\noindent\begin{minipage}{0.158\textwidth}\centering GT\end{minipage}%
\begin{minipage}{0.158\textwidth}\centering Dense\end{minipage}%
\begin{minipage}{0.158\textwidth}\centering 0.02 m\end{minipage}%
\begin{minipage}{0.158\textwidth}\centering 0.05 m\end{minipage}%
\begin{minipage}{0.158\textwidth}\centering 0.10 m\end{minipage}%
\begin{minipage}{0.158\textwidth}\centering 0.20 m\end{minipage}\par}
\vspace{1pt}
\includegraphics[width=\textwidth]{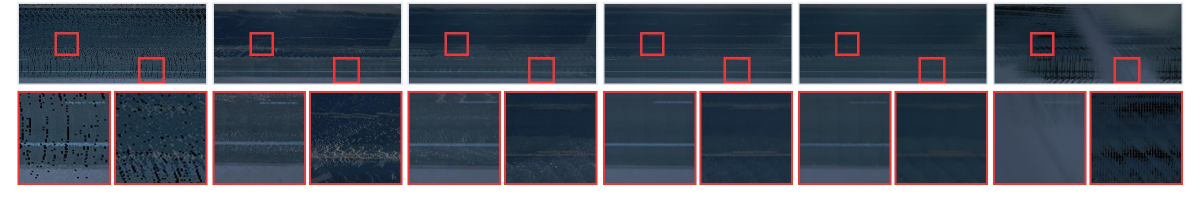}
\vspace{-5mm}
\caption{Qualitative comparison of dense prediction and grid fusion at different resolutions.}
\label{fig:grid-ablation-qualitative}
\vspace{-4mm}
\end{figure*}

Table~\ref{tab:quantitative-comparison}(a) reports the average results on
Waymo, and Table~\ref{tab:quantitative-comparison}(b) evaluates RoadVGGT on
nuScenes without training or fine-tuning on nuScenes. Across both datasets,
RoadVGGT achieves a better overall quantitative trade-off: it delivers better
reconstruction quality, maintains a compact representation, and preserves
comparable inference and rendering speed. In particular, compared with the
feed-forward AnySplat baseline, RoadVGGT is substantially more compact while
still delivering stronger overall reconstruction results. These results also demonstrate
the cross-dataset generalizability of RoadVGGT.
\subsection{Qualitative Comparison}
RoadVGGT produces RGB BEV, semantic BEV, elevation maps, and camera-view
renders from one compact representation. Figure~\ref{fig:qualitative}
compares AnySplat, RoGS, and RoadVGGT on Waymo and zero-shot nuScenes scenes.
The complete BEV panels show road appearance, semantics, and elevation,
while the enlarged regions and camera-view crop examine local appearance,
boundary artifacts, and lane-marking reconstruction. Compared with RoGS,
RoadVGGT retains more of the visible road surface because it filters
road-surface predictions using semantic masks rather than relying on a preset
lateral range around the driving trajectory, and it also produces clearer
markings, cleaner boundaries, and semantic and elevation maps that are closer
to the ground truth. Compared with AnySplat, RoadVGGT produces cleaner road
surfaces with fewer fused Gaussian artifacts and better organized local
structure. Overall, RoadVGGT achieves better qualitative reconstruction quality
across both datasets.
\subsection{Ablation Studies}
\subsubsection{Dense Prediction and Grid Resolution}
We first compare the dense pixel-wise representation with road-plane fusion
at four grid resolutions. Table~\ref{tab:ablation-summary}(a) and
Figure~\ref{fig:grid-ablation-qualitative} show that grid fusion greatly
reduces storage while suppressing noise in the dense predictions. Very fine
grids retain greater capacity for small structures, but are also more sensitive
to geometric noise; this finer-grained representation capability does not translate into
consistently better reconstruction and requires more storage.
Conversely, increasingly coarse grids suppress variation at the cost of local
road details and semantic structure. We therefore use \(0.05\) m as the default
balance between noise robustness, reconstruction quality, and compactness.
\subsubsection{2D versus 3D Gaussians}
\begin{figure}[t]
\centering
{\footnotesize
\noindent\begin{minipage}{0.326\columnwidth}\centering GT\end{minipage}%
\begin{minipage}{0.326\columnwidth}\centering 3D Gaussian\end{minipage}%
\begin{minipage}{0.326\columnwidth}\centering 2D Gaussian\end{minipage}\par}
\vspace{1pt}
\includegraphics[width=\columnwidth]{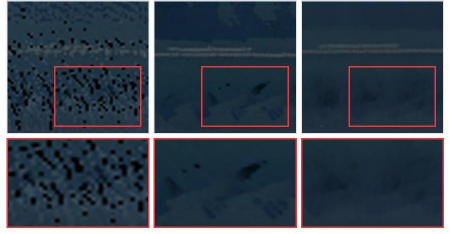}
\vspace{-5mm}
\caption{Qualitative comparison of 2D and 3D Gaussian parameterizations.}
\label{fig:2dg3dg-qual}
\vspace{-3mm}
\end{figure}

Road surfaces are locally close to two-dimensional manifolds, motivating our
use of flattened 2D Gaussians. We compare 2D and 3D Gaussian models that
differ only in their Gaussian parameterization.
Table~\ref{tab:ablation-summary}(b) shows that the 2D parameterization improves
overall reconstruction quality. Figure~\ref{fig:2dg3dg-qual} further shows that 2D Gaussians produce smoother
road surfaces and noticeably reduce noise in the BEV reconstruction.

\subsubsection{Road-Structure-Aware Grouping}
We compare the full model with variants without junction protection and
without both structure-aware components. Table~\ref{tab:ablation-summary}(c)
and Figure~\ref{fig:structure-qual} provide quantitative and camera-view
comparisons. The full design achieves the best quantitative
balance, while the qualitative comparisons show that removing both
components degrades local appearance and semantics. Removing junction
protection also worsens elevation accuracy.
\begin{center}
\begin{minipage}{0.98\columnwidth}
\centering
{\footnotesize
\noindent\begin{minipage}{0.326\columnwidth}\centering w/o Class+Junc.\end{minipage}%
\begin{minipage}{0.326\columnwidth}\centering w/o Junc.\end{minipage}%
\begin{minipage}{0.326\columnwidth}\centering Full\end{minipage}\par}
\vspace{1pt}
\includegraphics[width=\columnwidth]{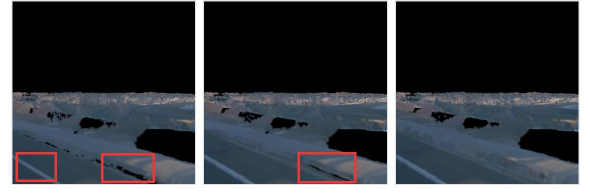}
\vspace{-5mm}
\captionof{figure}{Qualitative ablation of structure-aware grouping.}
\label{fig:structure-qual}
\vspace{0mm}
\end{minipage}
\end{center}

\section{Conclusion}
We presented RoadVGGT, a road-structure-aware feed-forward framework for
large-scale road surface reconstruction. By combining the geometric priors
of a pretrained geometric foundation model with a learned Gaussian head,
road-plane grid fusion, and structure-aware grouping, RoadVGGT converts
dense multi-view predictions into a compact road representation without
test-time scene-specific optimization.
Experiments show that RoadVGGT achieves strong BEV mapping quality together
with a compact direct-inference road representation. However, its performance still
depends heavily on the backbone's geometric estimation quality. Future work
will therefore focus on more accurate and robust geometric foundation models
for driving scenes to make downstream road reconstruction more reliable.
\bibliography{aaai2027}
\end{document}